%% file: main.tex
\documentclass[preprint,5p,times,twocolumn]{elsarticle}

\usepackage{amssymb}
\usepackage{amsmath,graphicx}
\usepackage{graphicx}
\usepackage{multirow}
\usepackage{flushend}
\usepackage[table,xcdraw]{xcolor}
\usepackage{comment}
\usepackage{xcolor}
\usepackage{xspace}

\newcommand{\mymethod}{{Gradient Artificial Distancing}\xspace}
\newcommand{\mymethodabbr}
{{\ensuremath{\mathrm{G
AD}}}\xspace}

\newcommand{\RS}{\mathit{RS}\xspace}
\newcommand{\RC}{\mathit{RC}\xspace}
\newcommand{\CH}{\mathit{CH}\xspace}
\newcommand{\revOne}[1]{\textcolor{black}{#1}}

\usepackage[linesnumbered,ruled,vlined]{algorithm2e}
\newcommand{\image}{\mathcal{I}\xspace}
\newcommand{\neuralnetwork}{\Xi\xspace}
\newcommand{\out}{\mathbf{out}\xspace}
\newcommand{\targets}{\mathbf{targets}\xspace}
\newcommand{\concatenate}{\mathit{concatenate}\xspace}
\newcommand{\explain}{\mathit{explain}\xspace}
\newcommand{\attr}{\mathbf{attr}\xspace}

\SetCommentSty{mycommfont}
\SetKwInput{KwInput}{Input}                %
\SetKwInput{KwOutput}{Output}              %

\usepackage{url}

\def\x{{\mathbf x}}

\journal{Pattern Recognition Letters}

\begin{document}

\begin{frontmatter}

\title{Transforming gradient-based techniques into interpretable methods}

\author[1,2,3]{Caroline Mazini Rodrigues}
\ead{caroline.mazinirodrigues@esiee.fr}
\author[1]{Nicolas Boutry}
\ead{nicolas.boutry@lrde.epita.fr}
\author[2]{Laurent Najman}
\ead{laurent.najman@esiee.fr}

\cortext[3]{Corresponding author.}

\affiliation[1]{organization={Laboratoire de Recherche de l'EPITA -- LRE},%
    addressline={14-16, Rue Voltaire}, 
    city={Le Kremlin-Bicêtre},
    postcode={94270}, 
    country={France}}
\affiliation[2]{organization={Univ Gustave Eiffel, CNRS, LIGM},%
    city={Marne-la-Vallée},
    postcode={77454}, 
    country={France}}

\begin{abstract}
The explication of Convolutional Neural Networks (CNN) through xAI techniques often poses challenges in interpretation. The inherent complexity of input features, notably pixels extracted from images, engenders complex correlations. Gradient-based methodologies, exemplified by Integrated Gradients (IG), effectively demonstrate the significance of these features. Nevertheless, the conversion of these explanations into images frequently yields considerable noise. Presently, we introduce \mymethodabbr (\mymethod) as a supportive framework for gradient-based techniques. Its primary objective is to accentuate influential regions by establishing distinctions between classes. The essence of \mymethodabbr is to limit the scope of analysis during visualization and, consequently reduce image noise. Empirical investigations involving occluded images have demonstrated that the identified regions through this methodology indeed play a pivotal role in facilitating class differentiation.
\end{abstract}

\begin{keyword}
explainable artificial intelligence \sep convolutional neural network \sep gradient-based \sep interpretability

\end{keyword}

\end{frontmatter}

\input{intro.tex}
\input{method.tex}
\input{experiments.tex}

\input{conclusion.tex}

\bibliographystyle{elsarticle-num}

\bibliography{paper}

\end{document}

%% file: intro.tex
\section{Introduction}

\textit{Post-hoc} and \textit{model-agnostic} explainable artificial intelligence (xAI) methods can be useful in explaining many types of general machine learning models. Within this group of methods, there are different types such as example-based explanations (\cite{erhan:tech:2009}, \cite{bien:statistics:2012},\cite{kim:nips:2016}), surrogate models (\cite{Ribeiro:CKDDM:2016},\cite{thiagarajan:arxiv:2016}), and influence methods (\cite{simonyan:iclr:2014}, \cite{merrick:corr:2019}, \cite{zeiler:eccv:2014}, \cite{bach:plosOne:2015}, \cite{lundberg:neurips:2017}). Of particular interest about the latter is that it tries to explain using the inner workings of the model being explained. Essentially, these explanations aim to reflect, to some extent, how the models think internally.

The challenge lies in the complexity of interpreting explanations derived from these methods, even when they are considered reliable. An instance of this difficulty arises with visual explanations that rely on attribution or salience maps, which can sometimes produce unclear or noisy representations. For instance, Figure~\ref{fig:exampleIG} demonstrates noisy visualizations generated by the Integrated Gradients (\cite{Sundararajan:2017:icml}) technique applied to a trained ResNet-18 model~\cite{He:cvpr:2016} aiming at providing localized explanations for a sample within a dataset. 

\revOne{Despite the presence of noise in their visualizations, these methods continue to be widely used across domains such as medicine~\cite{Borys:2023:ejr}, remote sensing~\cite{KAKOGEORGIOU:2021:ijaeog}, and transportation~\cite{kim:2022:plosone}. Even though these methods are good at capturing important features for the model, we must understand their visualizations to use them effectively and safely in these applications.}

\revOne{In our study, we present a method to simplify gradient-based visual explanations, aiming to make it easier to interpret CNN models' reasoning. This approach relies on the concept of class distancing.} The fundamental notion involves emphasizing the features crucial for distinguishing between two classes. By narrowing down the focus in this manner, our objective is to offer clearer and more streamlined visual explanations. Thus, we present the \mymethod (\mymethodabbr) technique, which employs a gradient-based method as the foundation of our iterative process. This methodology utilizes support regression models to reinforce the determination of important features. Each support model is trained to provide more distance between classes. Ultimately, we identify as important those features that consistently contribute to both the original model and the support models throughout this iterative process.

Additionally, \revOne{to assess the reliability and human interpretability of these simplifications, we introduce an evaluation methodology.} This methodology examines groups of significant pixels represented as regions (polygons), mimicking the way humans interpret these explanations. The experiments are structured into three sections: firstly, we conduct a quantitative assessment of the regions created by groups of significant pixels; secondly, we showcase the explanations derived from these clusters in the form of attribution maps; and finally, we conduct a comparative analysis between the regions generated by the original attribution maps and those generated by \mymethodabbr.

Our main contributions are:

\begin{enumerate}

    \item \revOne{The \mymethodabbr technique proposed to improve the interpretability of the explanations by simplifying gradient-based visualizations through the concept of class distancing analysis.}

    \item  A methodology inspired by human cognition, to assess attribution maps based on regions.
\end{enumerate}

In Section~\ref{sec:grad_methods}, we present gradient-based methods that can be used together with \mymethodabbr; Section~\ref{sec:intuition} presents the intuition of the method followed by the three steps of the method \mymethodabbr in Section~\ref{sec:method}; we describe metrics and the obtained results in Sections~\ref{sec:eval_metrics} and \ref{sec:results} respectively; and we conclude this paper presenting the future work in Section~\ref{sec:conclusion}.  

%% file: method.tex
\section{Gradient-based methods}
\label{sec:grad_methods}

\begin{figure*}[!ht]
\centering
\includegraphics[width=0.9\linewidth]{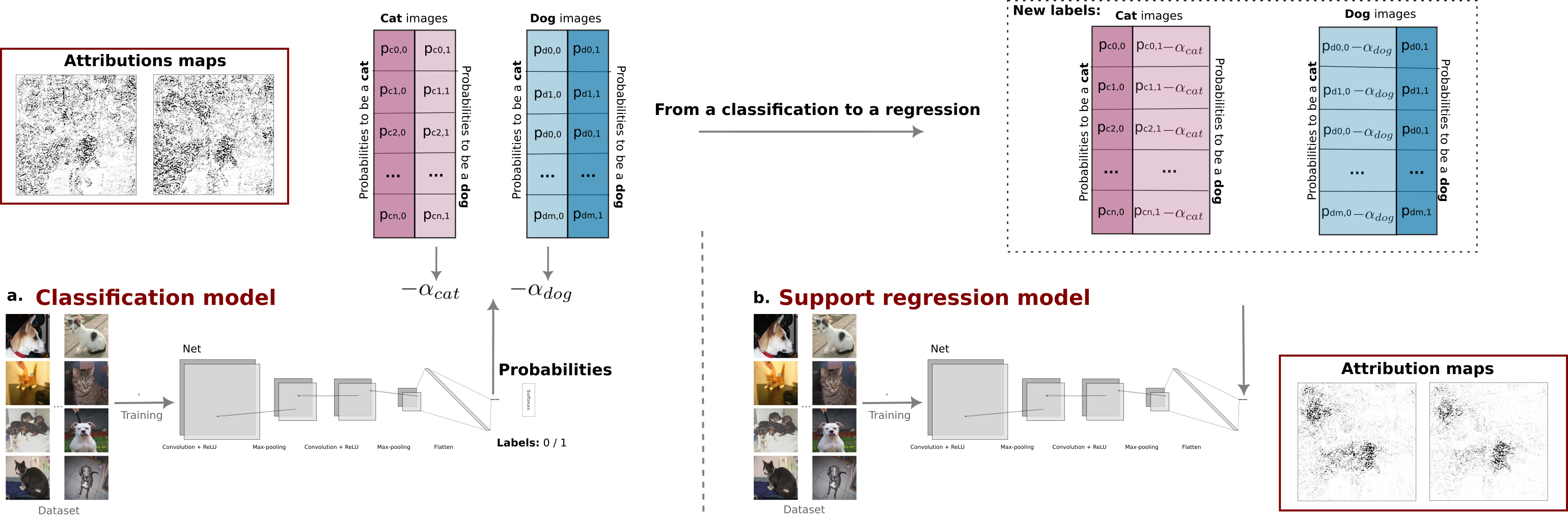}
\caption{Based on the non-normalized probabilities (before Softmax) for both classes we distance samples from different classes. We define values $\alpha_{cat}$ and $\alpha_{dog}$ that will be subtracted from the probabilities. The value $\alpha_{cat}$ will reduce the probability of cat images being dogs (light pink column) and $\alpha_{dog}$ will reduce the probability of dog images being cats (light blue column). This new probabilities are artificial; however, they preserve relations between samples from the same class.}
\label{fig:distance_classes}
\end{figure*}

This study targets a category of \textit{post-hoc} influence-based explanation techniques, the gradient-based methods. Influence methods aim to elucidate the learned model by presenting the impacts of inputs or internal components on the output~(\cite{adadi:IEEEAccess:2018}). These methods encompass sensitivity analysis with salience maps~\cite{simonyan:iclr:2014} or occlusion techniques~\cite{merrick:corr:2019},~\cite{zeiler:eccv:2014}; \textit{Layer-wise Relevance Propagation} (LRP)~\cite{bach:plosOne:2015}; and feature importance methods. For the purpose of this paper, we decided to concentrate on feature importance methods.

Our proposed methodology is founded on gradient-based methods. These methods fall within the category of feature importance methods, offering attribution maps that represent the importance of individual input features in relation to the output. They determine feature importance by manipulating the gradients of models. The fundamental principle behind these methods involves identifying the path that maximizes a specific output, thereby highlighting the most crucial input feature (the starting point of the path). Among the pioneering methods utilizing this approach is the Saliency visualization~\cite{simonyan:iclr:2014}, based on gradient ascent. Other notable techniques in this category encompass Deconvolution~\cite{zeiler:eccv:2014}, Gradients x Inputs~\cite{shrikumar:icml:2017}, Guided Backpropagation~\cite{Springenberg:ICLR:2015}, and Integrated Gradients (IG)~\cite{Sundararajan:2017:icml}.

\begin{figure}[!ht]
\centering
\includegraphics[width=6cm]{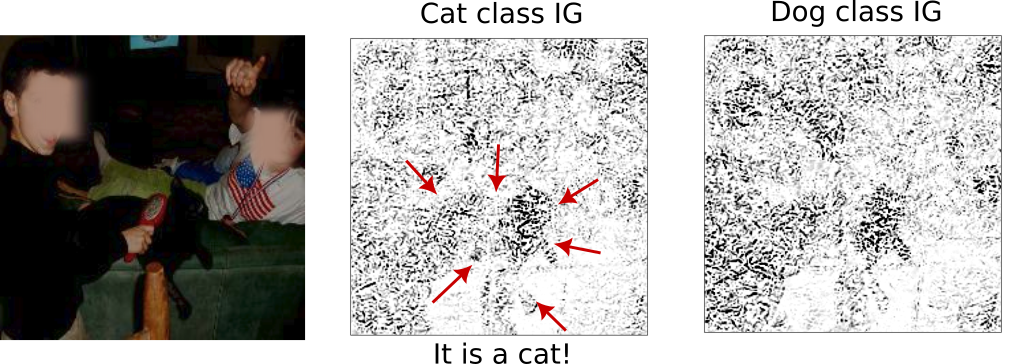}
\caption{Attribution maps obtained by Integrated Gradients method. The pixels' importance is described from white to black (less to more important) according to a chosen class.}
\label{fig:exampleIG}
\end{figure}

\revOne{Attributions, acquired at a pixel level, depend on the model's output, varying according to the examined class.} These attributions can be visualized collectively as an image, indicating the importance of pixels in determining class decisions. Figure~\ref{fig:exampleIG} shows the Integrated Gradients' method, depicting the most crucial pixels influencing the classification of \textit{cat} and \textit{dog}.

Despite outcome noise, as shown in Figure~\ref{fig:exampleIG}, these methods persist in widespread application, including medicine~\cite{Borys:2023:ejr}. Wang {\em et~al.}~\cite{wang:2023:neuroimage} conducted an evaluation of three xAI methods -- LRP, IG, and Guided Grad-CAM -- in the context of MRI Alzheimer's classification. They observed a substantial overlap among the three methods concerning brain regions, with IG showing the most promising results.

In fields like astronomy, gradient-based techniques are employed in research papers, such as the study conducted by Bhambra {\em et al.}~\cite{Bhambra:2022:mnras} concerning galaxy topology classification. They applied SmoothGRAD~\cite{smilkov:2017:smoothgrad} and found it to offer satisfactory explanations for their study.

Additionally, broader studies integrate gradient-based techniques in their investigations. Morrison {\em et al.}~\cite{morrison:2023:CVPR} utilize gradient methods to contrast attention-based models and CNN architectures with human perception. Woerl {\em et al.}~\cite{Woerl_2023_CVPR} analyze and improve the robustness of saliency maps for data-driven explanations.

\revOne{Given the prevalent application of these techniques, particularly in critical fields like medicine, we perceive an under-exploration concerning the interpretation of attribution map visualizations derived from gradient-based methods. Figure~\ref{fig:exampleIG} highlights the difficulty in understanding the model's reasoning solely through these visualizations. Despite their fidelity to the model's gradients and wide range of applications, these methods often do not produce clear and easily interpretable visualizations for humans.}

To address this problem, we introduce \mymethod (\mymethodabbr), designed to emphasize in the attribution maps solely those image regions that significantly contribute to class differentiation. This approach aims to minimize noise, thereby enhancing human interpretability.

\section{Intuition}
\label{sec:intuition}

Human perception tends to group nearby image pixels as a single entity, simplifying interpretation by reducing the number of components to analyze. Drawing clear boundaries around these components would ideally enhance the interpretability of an image region.

However, when the selected pixels lack evident proximity, defining components becomes more challenging. Consider Figure~\ref{fig:exampleIG} as an illustration, depicting important pixels in black within the second and third images. While there's a concentrated region of pixels that might represent the cat, the majority of the images display scattered black pixels, complicating interpretation.

We believe that we can enhance the interpretability of an explanation by reducing the scattered important pixels (depicted in black) to form smaller and denser regions. However, merely applying an intensity-based filter in the explanation image is not sufficient to maintain the network's accuracy. Such a filter would solely rely on an individual image each time, disregarding the network's comprehensive knowledge.

Our idea is based on a premise: the final activations' magnitudes in a network are negligible as long as, for the same set of images, the activations' order remains consistent. In essence, consider two networks providing activations $(1,-1), (0.5, -0.6), (-0.9, 0.8)$ for images $\mathbf{I}_1,\mathbf{I}_2$, and $\mathbf{I}_3$, respectively, exhibiting a similar order of activations for each class individually as another network with activations $(1,-2), (0.5, -1.6), (-1.9, 0.8)$. Both networks comprehend the three images similarly.

From the perspective of representation learning, this premise holds true as it maintains the spatial arrangement, preserving each image's position relative to its neighbors (akin to a simple translation in space). Nevertheless, the second set of activations exhibits a more pronounced distinction between the two classes.

\revOne{Our objective is to leverage this discrepancy between classes to minimize the number of analyzed image parts, emphasizing solely the most critical differences between classes, and reducing noise attributions, all while preserving fidelity to the original model.}

\section{Gradient artificial distancing}
\label{sec:method}

To find these most important class differences, we worked on the class activations from the training samples. Consider a classification problem with $m$ classes $ {\cal C} = \{c_d\}_{d=1}^{m}$, a dataset ${\cal D} = \bigcup \limits_{d=1}^{m} {\cal D}_{c_d}$, with $n_d$ samples ${\cal D}_{c_d} =  \{\mathbf{I}_{c_d,i}\}_{i=1}^{n_d}$ from the class $c_d$. Our goal is to accentuate class distinctions by artificially augmenting the disparity in final activations, effectively translating (distancing) the classes.

We aim to separate classes while preserving the original output space structure, maintaining the distribution of samples within the final activations' space. This strategy compels the network to focus solely on image regions crucial for creating the gap between classes, \revOne{also reducing possible noisy visualizations.}

With this method, we want to ensure that:

\begin{itemize}
   
    \item The updated activations' output space can be approximated using the identical network architecture and the original model's weights. Consequently, the optimization process aims to yield a close approximation with the initialization of the original weights.

    \item The final explanation is expected to have fewer highly important pixels, but it must not introduce new crucial pixels that were absent in the original explanation. Incorporating new important pixels would deviate the explanation from the knowledge embedded within the original model.
\end{itemize}

Three steps compose this process, as developed hereafter.

\subsection{Choosing classes of interest to increase distances}
\label{sec:step1}

We selected two specific classes from the set of trained classes, ${\cal C}$, with the aim of creating a distinct separation between them. For instance, consider the classes \textit{cat} and \textit{dog}. Our objective is to augment the separation by minimizing the likelihood of a dog being classified as a cat and vice-versa, all while maintaining the original learned structure.

We opted to modify the pre-Softmax activations instead of the post-Softmax ones to assess the increase in class separation, avoiding the normalization inherent in the Softmax function (which scales values between 0 and 1).

\revOne{Let us call the final non-normalized activations of a model for a dataset of $m$ classes and $n$ images as $\out$. Choosing two classes $c_k$ and $c_l$ we create a class artificial distancing by subtracting an $\alpha$ value from some activations in $\out$. The new matrix $\out'$ will have values $\out_{i,l}' = \out_{i,l} - \alpha_k$ for images $\mathbf{I}_i \in {\cal D}_{c_k}$ and $\out_{i,k}' = \mathit{P}_{i,k} - \alpha_l$ for images $\mathbf{I}_i \in {\cal D}_{c_l}$. We use this idea in Algorithm~\ref{alg:artificial} to generate the artificial distances for the next step of the methodology.}

\revOne{For example, if $c_k$ is the \textit{cat} class in an image dataset and $c_l$ the \textit{dog} class}, we directly alter the non-normalized activations of the likelihood of an image being a dog within ${\cal D}_{cat}$ (cat images) by subtracting a value $\alpha_{cat}$. Similarly, we subtract a value $\alpha_{dog}$ from the probabilities of ${\cal D}_{dog}$ samples (dog images) being classified as cats, depicted in Figure~\ref{fig:distance_classes}\textbf{a}. This process aims to diminish the consideration of cats as dogs and vice-versa, all artificially manipulated for this analysis.

\begin{algorithm}[!ht]
\DontPrintSemicolon
{
  \KwInput{a trained model $\neuralnetwork$; $\alpha_k$ and $\alpha_l$;}
  \KwData{Images $\mathbf{I}_{c_k,i}  \in {\cal D}_{c_k}$ from class $c_k$ and $\mathbf{I}_{c_l,j} \in {\cal D}_{c_l}$ from class $c_l$;}
  \KwOutput{Regression targets $\targets_{k,l}$ for images in ${\cal D}_{c_k}$ and ${\cal D}_{c_l}$.}

$\out_k := \neuralnetwork\left({\cal D}_{c_k}\right)$ \tcp*{outputs from images in class $c_k$}

$\out_l := \neuralnetwork\left({\cal D}_{c_l}\right)$ \tcp*{outputs from images in class $c_l$}

$\out_k(.,l) := \out_k(.,l) - \alpha_k$ \tcp*{reducing $l$ in class $k$}

$\out_l(.,k) := \out_l(.,k) - \alpha_l$ \tcp*{reducing $k$ in class $l$}

$\targets_{k,l} := \concatenate(\out_k, \out_l)$  
}
        
\caption{Artificial distance algorithm}
\label{alg:artificial}
\end{algorithm}

\subsection{Training regressions}
\label{sec:step2}

This artificial distancing process necessitates training new models to prompt the networks to generate these revised cat/dog output activations. These models are trained as regression problems, aiming to precisely obtain the altered artificial activations. While the inputs $x$ remain the images from ${\cal D}$ and the weights are based on the analyzed classification network for initialization, the expected outputs $y$ now represent the modified probabilities \revOne{$\targets$ obtained by Algorithm~\ref{alg:artificial}}. The primary goal is not to achieve more generalized models; rather, it's to attain the most accurate approximations of these new outputs, thereby discerning the persistently important features from the original explanation. To facilitate this, we replace the Softmax activation with a Linear one and employ Mean Squared Error (MSE) for regression training. Figure~\ref{fig:distance_classes}\textbf{b} illustrates this transformation in the model learning process, with labels converted into a probability vector.

\revOne{For every different pair of values $(\alpha_k,\alpha_l)_s$, we will generate new target values $\targets_s$ based on Algorithm~\ref{alg:artificial} to train a distinct model $\neuralnetwork_s$ for regression, as explained earlier. These models will serve as the support regression models.}

\subsection{Choosing important features}

\begin{figure}[!ht]
\centering
\includegraphics[width=\linewidth]{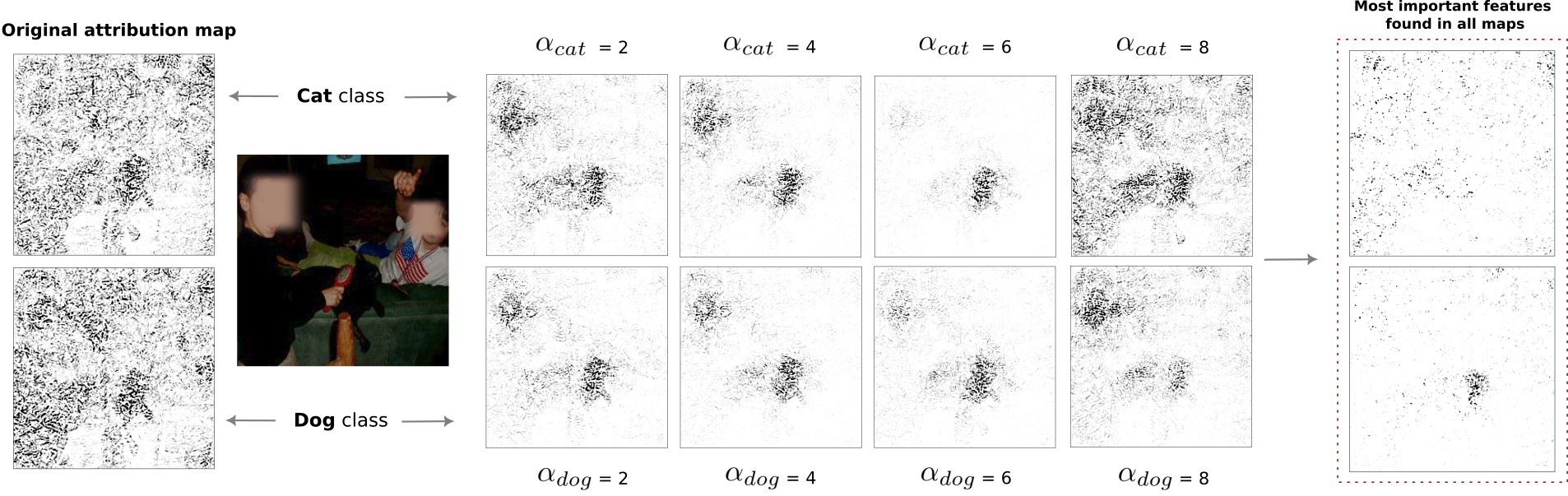}
\caption{Example of choosing important features. After training regression networks, we apply the IG method for each resulting model and both classes. The final attribution map includes only features present in all five attribution maps, the initial one (classification model analyzed), and the four support regression models. We obtain filtered attribution maps, including the regions of the image that most separate classes.}
\label{fig:example_combining}
\end{figure}

For precision in identifying crucial features, we propose iterating through steps 1 and 2 (Sections \ref{sec:step1} and \ref{sec:step2}) multiple times, progressively increasing the values of $\alpha_{k}$ and $\alpha_{l}$ \revOne{and training for each couple of $\alpha$ a support regression model}. Upon their training, each network is subject to the chosen Gradient-based xAI method when examining images. The significant features extracted should mirror those of the original model under explanation. However, we anticipate that the consistently important features across all networks will hold greater significance in differentiating the two classes. Figure \ref{fig:example_combining} illustrates the utilization of four support regression networks to delineate important regions. The analyzed image, in which Integrated Gradients (IG) was applied, was erroneously classified as a dog by the network. \revOne{We show the in Algorithm~\ref{alg:selection} how to combine the explanations of each support regression model to obtain a final attribution map $\attr$.}

\begin{algorithm}[!ht]
\DontPrintSemicolon
{
  \KwInput{Original model $\neuralnetwork$, set of trained support regression models $\{\neuralnetwork_0, ..., \neuralnetwork_s\}$, xAI gradient-based technique $\explain$;}
  \KwData{Image $\image_i$ to be explained;}
  \KwOutput{Attribution map $\attr$.}

 $\attr_{\mathit{orig}}:= \explain(\neuralnetwork, \image_i)$ \tcp*{original attribution map}
 
 $\attr_{\mathit{aux}}(\attr_{\mathit{orig}}>0):= 1 $ \tcp*{1's if positive attributions 0's otherwise}

 \tcc*{for each support regression network}
 \For{$\neuralnetwork_j \in \{\neuralnetwork_0, ..., \neuralnetwork_s\}$}{

       $\attr_j:= \explain(\neuralnetwork_j, \image_i)$  \tcp*{obtain attribution map}
       
       $\attr := \attr_j \times \attr_{\mathit{aux}}$ \tcp*{maintain attributions also in previous map}
       
       $\attr_{\mathit{aux}}(\attr>0):= 1$ \tcp*{update $\attr_{\mathit{aux}}$ as line 2}
    }
        
}
        
\caption{Features selection algorithm}
\label{alg:selection}
\end{algorithm}

\subsection{\revOne{Extension to a multi-class problem}}
\label{sec:multiclass}

\revOne{Until now, we've outlined the methodology applied to a two-class problem. In this section, we explore two approaches for adapting GAD to a multi-class problem.}

\revOne{\textbf{One vs. all (OvA):} With this strategy, we would compare one specific class, $c_k$, against all other classes. The adaptation involves selecting $c_k$ as the target class and treating all other images as belonging to a single class, $c_l$, in Algorithm~\ref{alg:artificial}. This approach would emphasize the distinct characteristics of $c_k$ while contrasting them with the features of all other classes.}

\revOne{\textbf{Split output space (Half):} With this strategy, we aim to create a more gradual effect on the output space compared to the \textbf{One vs. All} approach. The concept involves separating classes into two clusters based on their proximity in the output space, and then applying the artificial distancing (Algorithm~\ref{alg:artificial}) between these two clusters. For instance, in a four-class problem, suppose classes 0 and 2 form one cluster (corresponding to $c_k$), while classes 1 and 3 form another cluster (corresponding to $c_l$), based on their output similarities. We subtract $\alpha_k$ from the output of images belonging to classes 0 and 2 at positions 1 and 3 (to reduce the influence of $c_l$): $\out_k(.,[1,3]) := \out_k(.,[1,3]) - \alpha_k$ (line 3 of Algorithm~\ref{alg:artificial}). We do a similar process for classes 1 and 3: $\out_l(.,[0,2]) := \out_l(.,[0,2]) - \alpha_l$ (line 4 Algorithm~\ref{alg:artificial}).}

%% file: experiments.tex
\section{Metrics for evaluation}
\label{sec:eval_metrics}

We propose to evaluate our method using two criteria: \textit{Complexity} and \textit{Sensitivity}.

Regarding \textit{complexity}, our aim is to prioritize visualizations that are less complex, indicating reduced noise levels. In terms of \textit{sensitivity}, we seek to assess the consequences of occluding essential features on the model's output. We compare occluding important features identified by \mymethodabbr (our method) against occluding the remaining important features -- constituting a supplementary set found in the original explanation but absent in \mymethodabbr's explanation. The anticipation is that with \mymethodabbr, fewer important features will be identified, potentially yielding a more pronounced impact on the output.

\begin{figure}[!ht]
\centering
\includegraphics[width=0.7\linewidth]{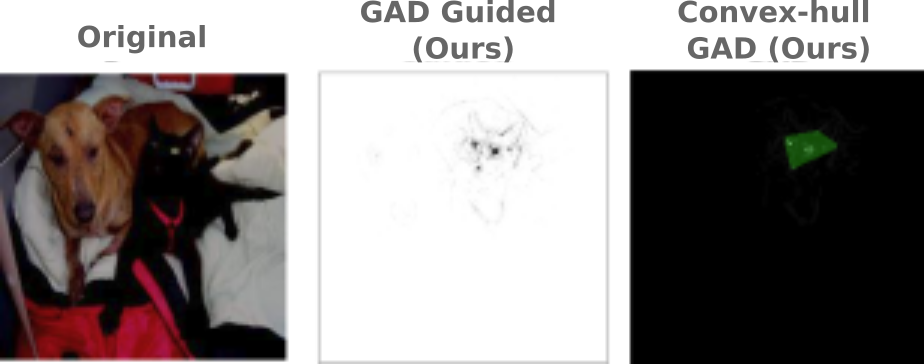}
\caption{Example of convex hull applied to an attribution map. We involve with the convex-hull the pixels reaching more than $10\%$ of the highest importance according to the GAD -- Guided Propagation attribution map.}
\label{fig:example_convex}
\end{figure}

As mentioned earlier, we aim to simulate human perception, which tends to group nearby pixels as one unit. To account for this in our evaluation, we create a convex hull involving a selected set of the most important pixels. Figure~\ref{fig:example_convex} illustrates an instance of the produced convex hull. Therefore, concerning \textit{complexity}, we assess the ratio $\RC$ by comparing the area of the convex hull in the original explanation ($\mathit{A}(\CH_{Orig})$) to the area of the convex hull in the \mymethodabbr explanation ($\mathit{A}(\CH_{\mymethodabbr})$):

\begin{equation}
    \RC = \frac{\mathit{A}(\CH_{\mymethodabbr})}{\mathit{A}(\CH_{Orig})}
    \label{eq:r_c}
\end{equation}

where $0 < \RC < 1.0$ represents a smaller important region for \mymethodabbr explanation. 

Regarding \textit{sensitivity}, we leverage the created convex hulls to conduct occlusions. These convex hulls, denoted as $CH_{Orig}$ and $CH_{\mymethodabbr}$, are represented in image masks $\mathit{M}_{Orig}$ and $\mathit{M}_{\mymethodabbr}$, respectively, with a value of 1. Internal or boundary pixels within the convex hulls are assigned a value of 1, while the remaining pixels receive 0 (forming a binary mask). Utilizing these masks, we generate two occlusions: one using $\mathit{M}_{\mymethodabbr}$ and another employing $\mathit{M}_{Orig} -\mathit{M}_{\mymethodabbr}$. Our anticipation is that the occlusion using $\mathit{M}_{\mymethodabbr}$ will exert a more substantial impact.

To quantify this impact, we propose calculating the ratio $\RS$, derived from the difference between the output before and after occlusion, divided by the area of the occluded section:

\begin{equation}
    \mathit{RS}(\mathbf{I},\mathit{M},a) = \frac{|N_a(\mathbf{I}) - N_a(\mathbf{I} \times (\mathbf{1} - \mathit{M}^{(a)}))|}{\sum_{i=0}^m \sum_{j=0}^n \mathit{M}_{i,j}^{(a)}}
    \label{eq:r_s}
\end{equation}

where $N_a(\mathbf{I})$ is the output $a$ of the network for an image $\mathbf{I} \in {\cal D}$ of size $m \times n$ and $\mathit{M}_{i,j}^{(a)}$ is the pixel $(i,j)$ from a mask of size $m \times n$ corresponding to the output $a$. We expect $\RS(\mathbf{I},\mathit{M}_{\mymethodabbr}, a) > \RS(\mathbf{I},\mathit{M}_{orig} - \mathit{M}_{\mymethodabbr}, a), \forall a \in {\cal C}$ (for all classes).

\section{Experiments and results}
\label{sec:results}

We experimented with two CNN architectures: VGG16~\cite{Simonyan:iclr:2015} and ResNet18~\cite{He:cvpr:2016}, utilizing two distinct datasets. The first dataset involves a binary classification of cats and dogs~\footnote{https://www.kaggle.com/competitions/dogs-vs-cats-redux-kernels-edition/data}, while the second dataset, the CUB-200-2011 dataset~\cite{he:2019:icsvt}, focuses on bird classification. For the bird classification task, we specifically chose the \textit{warbler} and \textit{sparrow} species as the two classes (the original dataset comprises 20 classes each, which we put together for this analysis).

We applied values $\alpha = {0, 2, 4, 6, 8}$ for both classes. The support regression networks initiated the training using weights from the analyzed classification model (original model). Each support model underwent training for 10 epochs, utilizing the Adam optimizer and a learning rate set to $4\times 10^{-5}$.

\revOne{We conducted experiments applying \mymethodabbr to five explanation techniques commonly used in the literature: Saliency~\cite{simonyan:iclr:2014}, Deconvolution~\cite{zeiler:eccv:2014}, Gradient x Input~\cite{shrikumar:icml:2017}, Guided-Backpropagation~\cite{Springenberg:ICLR:2015}, and Integrated Gradients~\cite{Sundararajan:2017:icml}. These techniques served as the xAI technique $\explain$ in Algorithm~\ref{alg:selection}. We present the results of \mymethodabbr alongside the original explanations produced by each xAI technique for comparison.}

In the subsequent experiments on \textit{complexity} and \textit{sensitivity}, we employed a total of 512 images (for both datasets), evenly distributed with 256 images per class. To create the convex hull, we filtered out $50\%$ of the pixels, retaining only the most important half.

\subsection{Complexity}

Following the computation of $\RC$ (Equation~\ref{eq:r_c}) for all 512 images, we count the number of images where $\RC<1.0$, indicating a point in favor of \mymethodabbr, and those where $\RC \geq 1.0$, indicating a point in favor of the original visualization.

\begin{table}[!ht]
\centering
\caption{For most methods, \mymethodabbr is able to reduce interest area in visualizations. Evaluation of $\RC$ for five literature methods, Saliency (S), Deconvolution (D), Gradient x Input (GxI), Guided-Backpropagation (GB) and Integrated Gradients (IG); two architectures, ResNet18 and VGG16; and two datasets, Cat x Dog and Birds. Orig.1 and Orig.2 represent the original methods' visualization for the two classes. \mymethodabbr 1 and \mymethodabbr 2 are the application of our method to both classes visualizations.}
\label{tab:complexity}
\resizebox{.4 \textwidth}{!}{
\begin{tabular}{cccccccc}
\textbf{}                                                                                             & \textbf{}                                                                      & \textbf{}                              & \textbf{S}                                                  & \textbf{D}                                                  & \textbf{GxI}                                                & \textbf{GB}                                                 & \textbf{IG}                                                 \\ \hline
\multicolumn{1}{c|}{}                                                                                 & \multicolumn{1}{c|}{}                                                          & \textbf{Orig.1}                         & {\color[HTML]{CB0000} \textbf{357}}                         & 0                                                           & {\color[HTML]{333333} 15}                                   & {\color[HTML]{333333} 4}                                    & {\color[HTML]{333333} 17}                                   \\
\multicolumn{1}{c|}{}                                                                                 & \multicolumn{1}{c|}{}                                                          & \textbf{\mymethodabbr 1 (ours)}                         & {\color[HTML]{333333} 155}                                  & {\color[HTML]{00009B} \textbf{512}}                         & {\color[HTML]{00009B} \textbf{490}}                         & {\color[HTML]{00009B} \textbf{427}}                         & {\color[HTML]{00009B} \textbf{480}}                         \\
\multicolumn{1}{c|}{}                                                                                 & \multicolumn{1}{c|}{}                                                          & \cellcolor[HTML]{DAE8FC}\textbf{Orig.2} & \cellcolor[HTML]{DAE8FC}159                                 & \cellcolor[HTML]{DAE8FC}0                                   & \cellcolor[HTML]{DAE8FC}{\color[HTML]{333333} 2}            & \cellcolor[HTML]{DAE8FC}{\color[HTML]{333333} 3}            & \cellcolor[HTML]{DAE8FC}{\color[HTML]{333333} 0}            \\
\multicolumn{1}{c|}{}                                                                                 & \multicolumn{1}{c|}{\multirow{-4}{*}{\textbf{VGG}}}                            & \cellcolor[HTML]{DAE8FC}\textbf{\mymethodabbr 2 (ours)} & \cellcolor[HTML]{DAE8FC}{\color[HTML]{00009B} \textbf{348}} & \cellcolor[HTML]{DAE8FC}{\color[HTML]{00009B} \textbf{512}} & \cellcolor[HTML]{DAE8FC}{\color[HTML]{00009B} \textbf{494}} & \cellcolor[HTML]{DAE8FC}{\color[HTML]{00009B} \textbf{428}} & \cellcolor[HTML]{DAE8FC}{\color[HTML]{00009B} \textbf{486}} \\ \cline{2-8} 
\multicolumn{1}{c|}{}                                                                                 & \multicolumn{1}{c|}{\cellcolor[HTML]{FFFFC7}}                                  & \textbf{Orig.1}                         & 122                                                         & 0                                                           & {\color[HTML]{333333} 2}                                    & {\color[HTML]{333333} 10}                                   & {\color[HTML]{333333} 7}                                    \\
\multicolumn{1}{c|}{}                                                                                 & \multicolumn{1}{c|}{\cellcolor[HTML]{FFFFC7}}                                  & \textbf{\mymethodabbr 1 (ours)}                         & {\color[HTML]{00009B} \textbf{390}}                         & {\color[HTML]{00009B} \textbf{512}}                         & {\color[HTML]{00009B} \textbf{492}}                         & {\color[HTML]{00009B} \textbf{453}}                         & {\color[HTML]{00009B} \textbf{497}}                         \\
\multicolumn{1}{c|}{}                                                                                 & \multicolumn{1}{c|}{\cellcolor[HTML]{FFFFC7}}                                  & \cellcolor[HTML]{DAE8FC}\textbf{Orig.2} & \cellcolor[HTML]{DAE8FC}130                                 & \cellcolor[HTML]{DAE8FC}0                                   & \cellcolor[HTML]{DAE8FC}{\color[HTML]{333333} 3}            & \cellcolor[HTML]{DAE8FC}{\color[HTML]{333333} 5}            & \cellcolor[HTML]{DAE8FC}{\color[HTML]{333333} 2}            \\
\multicolumn{1}{c|}{\multirow{-8}{*}{\textbf{\begin{tabular}[c]{@{}c@{}}Cat\\ x\\ Dog\end{tabular}}}} & \multicolumn{1}{c|}{\multirow{-4}{*}{\cellcolor[HTML]{FFFFC7}\textbf{ResNet}}} & \cellcolor[HTML]{DAE8FC}\textbf{\mymethodabbr 2 (ours)} & \cellcolor[HTML]{DAE8FC}{\color[HTML]{00009B} \textbf{380}} & \cellcolor[HTML]{DAE8FC}{\color[HTML]{00009B} \textbf{512}} & \cellcolor[HTML]{DAE8FC}{\color[HTML]{00009B} \textbf{490}} & \cellcolor[HTML]{DAE8FC}{\color[HTML]{00009B} \textbf{460}} & \cellcolor[HTML]{DAE8FC}{\color[HTML]{00009B} \textbf{488}} \\ \hline
\multicolumn{1}{c|}{}                                                                                 & \multicolumn{1}{c|}{}                                                          & \textbf{Orig.1}                         & {\color[HTML]{CB0000} \textbf{270}}                         & 0                                                           & {\color[HTML]{333333} 9}                                    & {\color[HTML]{333333} 0}                                    & {\color[HTML]{333333} 1}                                    \\
\multicolumn{1}{c|}{}                                                                                 & \multicolumn{1}{c|}{}                                                          & \textbf{\mymethodabbr 1 (ours)}                         & {\color[HTML]{333333} 240}                                  & {\color[HTML]{00009B} \textbf{511}}                         & {\color[HTML]{00009B} \textbf{488}}                         & {\color[HTML]{00009B} \textbf{450}}                         & {\color[HTML]{00009B} \textbf{500}}                         \\
\multicolumn{1}{c|}{}                                                                                 & \multicolumn{1}{c|}{}                                                          & \cellcolor[HTML]{DAE8FC}\textbf{Orig.2} & \cellcolor[HTML]{DAE8FC}{\color[HTML]{CB0000} \textbf{301}} & \cellcolor[HTML]{DAE8FC}0                                   & \cellcolor[HTML]{DAE8FC}10                                  & \cellcolor[HTML]{DAE8FC}{\color[HTML]{333333} 0}            & \cellcolor[HTML]{DAE8FC}{\color[HTML]{333333} 0}            \\
\multicolumn{1}{c|}{}                                                                                 & \multicolumn{1}{c|}{\multirow{-4}{*}{\textbf{VGG}}}                            & \cellcolor[HTML]{DAE8FC}\textbf{\mymethodabbr 2 (ours)} & \cellcolor[HTML]{DAE8FC}{\color[HTML]{333333} 210}          & \cellcolor[HTML]{DAE8FC}{\color[HTML]{00009B} \textbf{510}} & \cellcolor[HTML]{DAE8FC}{\color[HTML]{00009B} \textbf{494}} & \cellcolor[HTML]{DAE8FC}{\color[HTML]{00009B} \textbf{451}} & \cellcolor[HTML]{DAE8FC}{\color[HTML]{00009B} \textbf{504}} \\ \cline{2-8} 
\multicolumn{1}{c|}{}                                                                                 & \multicolumn{1}{c|}{\cellcolor[HTML]{FFFFC7}}                                  & \textbf{Orig.1}                         & 153                                                         & 0                                                           & 1                                                           & {\color[HTML]{333333} 0}                                    & {\color[HTML]{333333} 0}                                    \\
\multicolumn{1}{c|}{}                                                                                 & \multicolumn{1}{c|}{\cellcolor[HTML]{FFFFC7}}                                  & \textbf{\mymethodabbr 1 (ours)}                         & {\color[HTML]{00009B} \textbf{358}}                         & {\color[HTML]{00009B} \textbf{512}}                         & {\color[HTML]{00009B} \textbf{504}}                         & {\color[HTML]{00009B} \textbf{472}}                         & {\color[HTML]{00009B} \textbf{506}}                         \\
\multicolumn{1}{c|}{}                                                                                 & \multicolumn{1}{c|}{\cellcolor[HTML]{FFFFC7}}                                  & \cellcolor[HTML]{DAE8FC}\textbf{Orig.2} & \cellcolor[HTML]{DAE8FC}174                                 & \cellcolor[HTML]{DAE8FC}0                                   & \cellcolor[HTML]{DAE8FC}{\color[HTML]{333333} 2}            & \cellcolor[HTML]{DAE8FC}{\color[HTML]{333333} 0}            & \cellcolor[HTML]{DAE8FC}{\color[HTML]{333333} 0}            \\
\multicolumn{1}{c|}{\multirow{-8}{*}{\textbf{Birds}}}                                                 & \multicolumn{1}{c|}{\multirow{-4}{*}{\cellcolor[HTML]{FFFFC7}\textbf{ResNet}}} & \cellcolor[HTML]{DAE8FC}\textbf{\mymethodabbr 2 (ours)} & \cellcolor[HTML]{DAE8FC}{\color[HTML]{00009B} \textbf{337}} & \cellcolor[HTML]{DAE8FC}{\color[HTML]{00009B} \textbf{512}} & \cellcolor[HTML]{DAE8FC}{\color[HTML]{00009B} \textbf{506}} & \cellcolor[HTML]{DAE8FC}{\color[HTML]{00009B} \textbf{470}} & \cellcolor[HTML]{DAE8FC}{\color[HTML]{00009B} \textbf{504}}
\end{tabular}
}
\end{table}

We show, in Table~\ref{tab:complexity}, the number of images where \mymethodabbr (\textbf{Ours}) and the original method exhibited smaller complexity (smaller important areas). Through the experiments involving five literature methods -- Saliency (S), Deconvolution (D), Gradient x Input (GxI), Guided-Backpropagation (GB), and Integrated Gradients (IG) -- we observed that \mymethodabbr successfully reduces the area of importance for nearly all images and methods.

\subsection{Sensitivity}

Upon computing $\RS(\mathbf{I},\mathit{M},a)$ (Equation~\ref{eq:r_s}) for the four occlusion masks, which include two masks per class $a$ ($\mathit{M}_{\mymethodabbr}$ and $\mathit{M}_{Orig} -\mathit{M}_{\mymethodabbr}$), we acquire four respective $\RS$ values each time. We designated the mask $\mathit{M}_{Orig} -\mathit{M}_{\mymethodabbr}$ as the \textit{supplementary} mask, denoted as \textbf{Sup.}, and $\mathit{M}_{\mymethodabbr}$ simply as \textbf{\mymethodabbr}. Therefore, the resulting four values correspond to: $\RS(\mathbf{I},\mymethodabbr_1,1)$, $\RS(\mathbf{I},\mymethodabbr_2,2)$, $\RS(\mathbf{I},\mathit{Sup.}_1,1)$, and $\RS(\mathbf{I},\mathit{Sup.}_2,2)$.

\begin{table}[!ht]
\centering
\caption{Four out of five literature methods indicate a bigger impact when occluding the important regions found by \mymethodabbr. %
Evaluation of $\RS$ for five literature methods, Saliency (S), Deconvolution (D), Gradient x Input (GxI), Guided-Backpropagation (GB) and Integrated Gradients (IG); two architectures, ResNet18 and VGG16; and two datasets, Cat x Dog and Birds. Sup.1, Sup.2, \mymethodabbr 1, and \mymethodabbr 2 indicate the average $\RS$ values using these for different masks for occlusion. Sup.1 and Sup.2 represent the supplementary masks for the two classes. \mymethodabbr 1 and \mymethodabbr 2 are our method masks for both classes. Values in $10^{2}$ scale.}
\label{tab:sensitivity_all}
\resizebox{.45 \textwidth}{!}{
\begin{tabular}{cccccccc}
\textbf{}                                                                                             & \textbf{}                                 &              & \textbf{S}                & \textbf{D}                     & \textbf{GxI}                         & \textbf{GB}                                                    & \textbf{IG}                                                    \\ \hline
\multicolumn{1}{c|}{}                                                                                 & \multicolumn{1}{c|}{}                                                          & \textbf{Sup.1}                         & 0.0800                                                         & 0.0029                         & 0.0052                                                         & 0.1100                                                         & 0.0100                                                         \\
\multicolumn{1}{c|}{}                                                                                 & \multicolumn{1}{c|}{}                                                          & \textbf{\mymethodabbr 1 (ours)}                         & {\color[HTML]{00009B} \textbf{0.1300}}                         & {\color[HTML]{00009B} \textbf{0.0056}}                         & {\color[HTML]{00009B} \textbf{0.1200}}                         & {\color[HTML]{00009B} \textbf{1.5900}}                         & {\color[HTML]{00009B} \textbf{0.1300}}                         \\
\multicolumn{1}{c|}{}                                                                                 & \multicolumn{1}{c|}{}                                                          & \cellcolor[HTML]{DAE8FC}\textbf{Sup.2} & \cellcolor[HTML]{DAE8FC}0.2400                                 & \cellcolor[HTML]{DAE8FC}0.0029 & \cellcolor[HTML]{DAE8FC}0.0051                                 & \cellcolor[HTML]{DAE8FC}0.1500                                 & \cellcolor[HTML]{DAE8FC}0.0096                                 \\
\multicolumn{1}{c|}{}                                                                                 & \multicolumn{1}{c|}{\multirow{-4}{*}{\textbf{VGG}}}                            & \cellcolor[HTML]{DAE8FC}\textbf{\mymethodabbr 2 (ours)} & \cellcolor[HTML]{DAE8FC}{\color[HTML]{00009B} \textbf{0.2500}} & \cellcolor[HTML]{DAE8FC} {\color[HTML]{00009B} \textbf{0.0062}} & \cellcolor[HTML]{DAE8FC}{\color[HTML]{00009B} \textbf{0.5600}} & \cellcolor[HTML]{DAE8FC}{\color[HTML]{00009B} \textbf{1.8200}} & \cellcolor[HTML]{DAE8FC}{\color[HTML]{00009B} \textbf{0.4900}} \\ \cline{2-8} 
\multicolumn{1}{c|}{}                                                                                 & \multicolumn{1}{c|}{\cellcolor[HTML]{FFFFC7}}                                  & \textbf{Sup.1}                         & {\color[HTML]{CB0000} \textbf{0.0600}}                         & 0.0013                         & 0.0016                                                         & 0.1300                                       \textbf{  }                & 0.0016                                                         \\
\multicolumn{1}{c|}{}                                                                                 & \multicolumn{1}{c|}{\cellcolor[HTML]{FFFFC7}}                                  & \textbf{\mymethodabbr 1 (ours)}                         & 0.0300                                                         & {\color[HTML]{00009B} \textbf{0.0024}}                         & {\color[HTML]{00009B} \textbf{0.0700}}                         & {\color[HTML]{00009B} \textbf{0.5100}}                         & {\color[HTML]{00009B} \textbf{0.1200}}                         \\
\multicolumn{1}{c|}{}                                                                                 & \multicolumn{1}{c|}{\cellcolor[HTML]{FFFFC7}}                                  & \cellcolor[HTML]{DAE8FC}\textbf{Sup.2} & \cellcolor[HTML]{DAE8FC}0.0500                                 & \cellcolor[HTML]{DAE8FC}0.0012 & \cellcolor[HTML]{DAE8FC}0.0016                                 & \cellcolor[HTML]{DAE8FC}0.1000                                 & \cellcolor[HTML]{DAE8FC}0.0020                                 \\
\multicolumn{1}{c|}{\multirow{-8}{*}{\textbf{\begin{tabular}[c]{@{}c@{}}Cat\\ x\\ Dog\end{tabular}}}} & \multicolumn{1}{c|}{\multirow{-4}{*}{\cellcolor[HTML]{FFFFC7}\textbf{ResNet}}} & \cellcolor[HTML]{DAE8FC}\textbf{\mymethodabbr 2 (ours)} & \cellcolor[HTML]{DAE8FC}{\color[HTML]{00009B} \textbf{0.0700}} & \cellcolor[HTML]{DAE8FC}{\color[HTML]{00009B} \textbf{0.0024}} & \cellcolor[HTML]{DAE8FC}{\color[HTML]{00009B} \textbf{0.1600}} & \cellcolor[HTML]{DAE8FC}{\color[HTML]{00009B} \textbf{0.4700}} & \cellcolor[HTML]{DAE8FC}{\color[HTML]{00009B} \textbf{0.1600}} \\ \hline
\multicolumn{1}{c|}{}                                                                                 & \multicolumn{1}{c|}{}                                                          & \textbf{Sup.1}                         & {\color[HTML]{CB0000} \textbf{0.2500}}                         & 0.0027                         & 0.0800                                                         & 0.0100                                                         & 0.0100                                                         \\
\multicolumn{1}{c|}{}                                                                                 & \multicolumn{1}{c|}{}                                                          & \textbf{\mymethodabbr 1 (ours)}                         & 0.2000                                                         &{\color[HTML]{00009B} \textbf{0.0056}}                         & {\color[HTML]{00009B} \textbf{0.5400}}                         & {\color[HTML]{00009B} \textbf{1.2800}}                         & {\color[HTML]{00009B} \textbf{0.5500}}                         \\
\multicolumn{1}{c|}{}                                                                                 & \multicolumn{1}{c|}{}                                                          & \cellcolor[HTML]{DAE8FC}\textbf{Sup.2} & \cellcolor[HTML]{DAE8FC}{\color[HTML]{CB0000} \textbf{0.1500}} & \cellcolor[HTML]{DAE8FC}0.0030 & \cellcolor[HTML]{DAE8FC}0.0092                                 & \cellcolor[HTML]{DAE8FC}0.0100                                 & \cellcolor[HTML]{DAE8FC}0.0072                                 \\
\multicolumn{1}{c|}{}                                                                                 & \multicolumn{1}{c|}{\multirow{-4}{*}{\textbf{VGG}}}                            & \cellcolor[HTML]{DAE8FC}\textbf{\mymethodabbr 2 (ours)} & \cellcolor[HTML]{DAE8FC}0.1300                                 & \cellcolor[HTML]{DAE8FC}{\color[HTML]{00009B} \textbf{0.0068}} & \cellcolor[HTML]{DAE8FC}{\color[HTML]{00009B} \textbf{0.3600}} & \cellcolor[HTML]{DAE8FC}{\color[HTML]{00009B} \textbf{1.4100}} & \cellcolor[HTML]{DAE8FC}{\color[HTML]{00009B} \textbf{0.3800}} \\ \cline{2-8} 
\multicolumn{1}{c|}{}                                                                                 & \multicolumn{1}{c|}{\cellcolor[HTML]{FFFFC7}}                                  & \textbf{Sup.1}                         & {\color[HTML]{CB0000} \textbf{0.1100}}                         & 0.0020                         & 0.0045                                                         & 0.0100                                                         & 0.0094                                                         \\
\multicolumn{1}{c|}{}                                                                                 & \multicolumn{1}{c|}{\cellcolor[HTML]{FFFFC7}}                                  & \textbf{\mymethodabbr 1 (ours)}                         & 0.0400                                                         & {\color[HTML]{00009B} \textbf{0.0048}}                         & {\color[HTML]{00009B} \textbf{0.1700}}                         & {\color[HTML]{00009B} \textbf{1.0300}}                         & {\color[HTML]{00009B} \textbf{0.2900}}                         \\
\multicolumn{1}{c|}{}                                                                                 & \multicolumn{1}{c|}{\cellcolor[HTML]{FFFFC7}}                                  & \cellcolor[HTML]{DAE8FC}\textbf{Sup.2} & \cellcolor[HTML]{DAE8FC}0.0500                                 & \cellcolor[HTML]{DAE8FC}0.0019 & \cellcolor[HTML]{DAE8FC}0.0037                                 & \cellcolor[HTML]{DAE8FC}0.0100                                 & \cellcolor[HTML]{DAE8FC}0.0026                                 \\
\multicolumn{1}{c|}{\multirow{-8}{*}{\textbf{Birds}}}                                                 & \multicolumn{1}{c|}{\multirow{-4}{*}{\cellcolor[HTML]{FFFFC7}\textbf{ResNet}}} & \cellcolor[HTML]{DAE8FC}\textbf{\mymethodabbr 2 (ours)} & \cellcolor[HTML]{DAE8FC}0.0500                                 & \cellcolor[HTML]{DAE8FC}{\color[HTML]{00009B} \textbf{0.0048}} & \cellcolor[HTML]{DAE8FC}{\color[HTML]{00009B} \textbf{0.1600}} & \cellcolor[HTML]{DAE8FC}{\color[HTML]{00009B} \textbf{1.0400}} & \cellcolor[HTML]{DAE8FC}{\color[HTML]{00009B} \textbf{0.2000}}
\end{tabular}
}
\end{table}

We show, in Table~\ref{tab:sensitivity_all}, the average $\RS$ values across all 512 images for \mymethodabbr 1, \mymethodabbr 2, Sup.1, and Sup.2, for the five literature methods, two architectures, and two datasets. For Deconvolution, Gradient x Input, Guided-Backpropagation, and Integrated Gradients, our method (\mymethodabbr) effectively identifies the most impactful regions for each decision, excluding the supplementary area (from the original visualization) that does not exhibit a greater impact (per pixel). However, in the case of Saliency visualization, both the areas selected by \mymethodabbr and the supplementary areas display similar impacts, indicating limited benefits of our method for this particular technique.

\subsection{\mymethodabbr attribution maps}
\begin{figure*}[!ht]
\centering
 \includegraphics[width=0.68\textwidth]{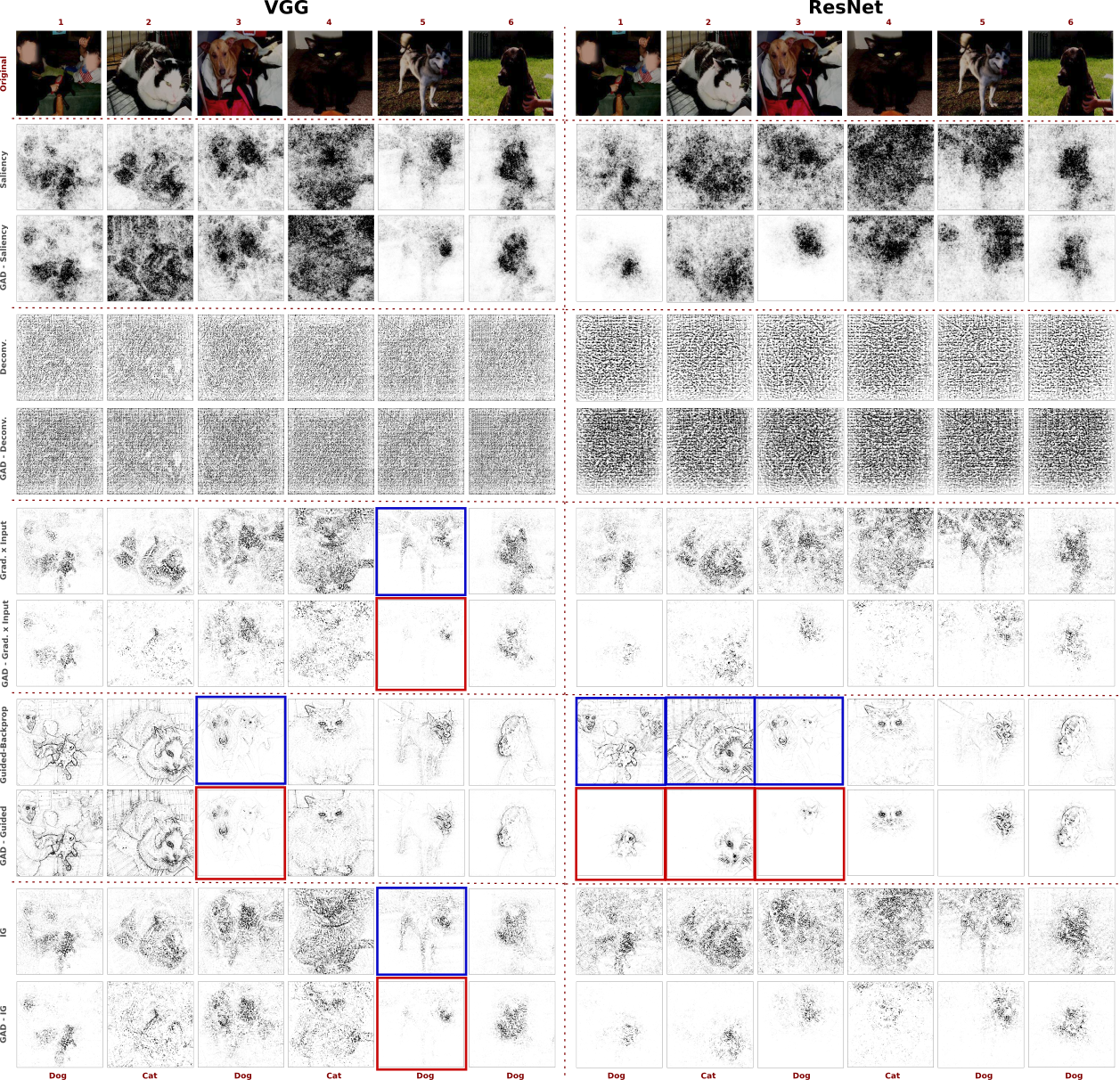} 
\caption{ \textbf{Our final visualizations:} \mymethodabbr improves interpretability of attribution maps. We present two-by-two rows of original attribution maps and \mymethodabbr maps from five gradient-based techniques: Saliency, Deconvolution, Gradient x Input, Guided-Backpropagation, and Integrated Gradients. We present on the right side the VGG results and on the left side the ResNet ones. At the bottom, we present the obtained classification for each image (cat or dog).}
\label{fig:cat_dog_attributions}
\end{figure*}
We present in Figure~\ref{fig:cat_dog_attributions} the qualitative results of employing \mymethodabbr with five gradient-based techniques. Each set of two rows illustrates the original attribution maps in the first row and the corresponding \mymethodabbr maps in the second row. These pairs of rows represent the tested methods: Saliency, Deconvolution, Gradient x Input, Guided-Backpropagation, and Integrated Gradients. The results for VGG are displayed on the right side, while those for ResNet are on the left. The classification results (cat or dog) are presented at the bottom. Notable examples highlight the enhancements in visualizations: the original attribution maps are represented in blue, and the corresponding \mymethodabbr visualizations are shown in red. For instance, Image 3 (highlighted in both networks) demonstrates a noteworthy distinction: both VGG and ResNet original attribution maps (in blue) appear similar in identifying the image as a dog. However, upon applying \mymethodabbr, distinct attribution maps (in red) emerge, with VGG focusing on the dog while ResNet emphasizes the cat, despite both networks classifying the image as a dog.

\subsection{Finding visual knowledge clues}

\begin{figure*}[!h]
\centering
 \includegraphics[width=0.45\textwidth]{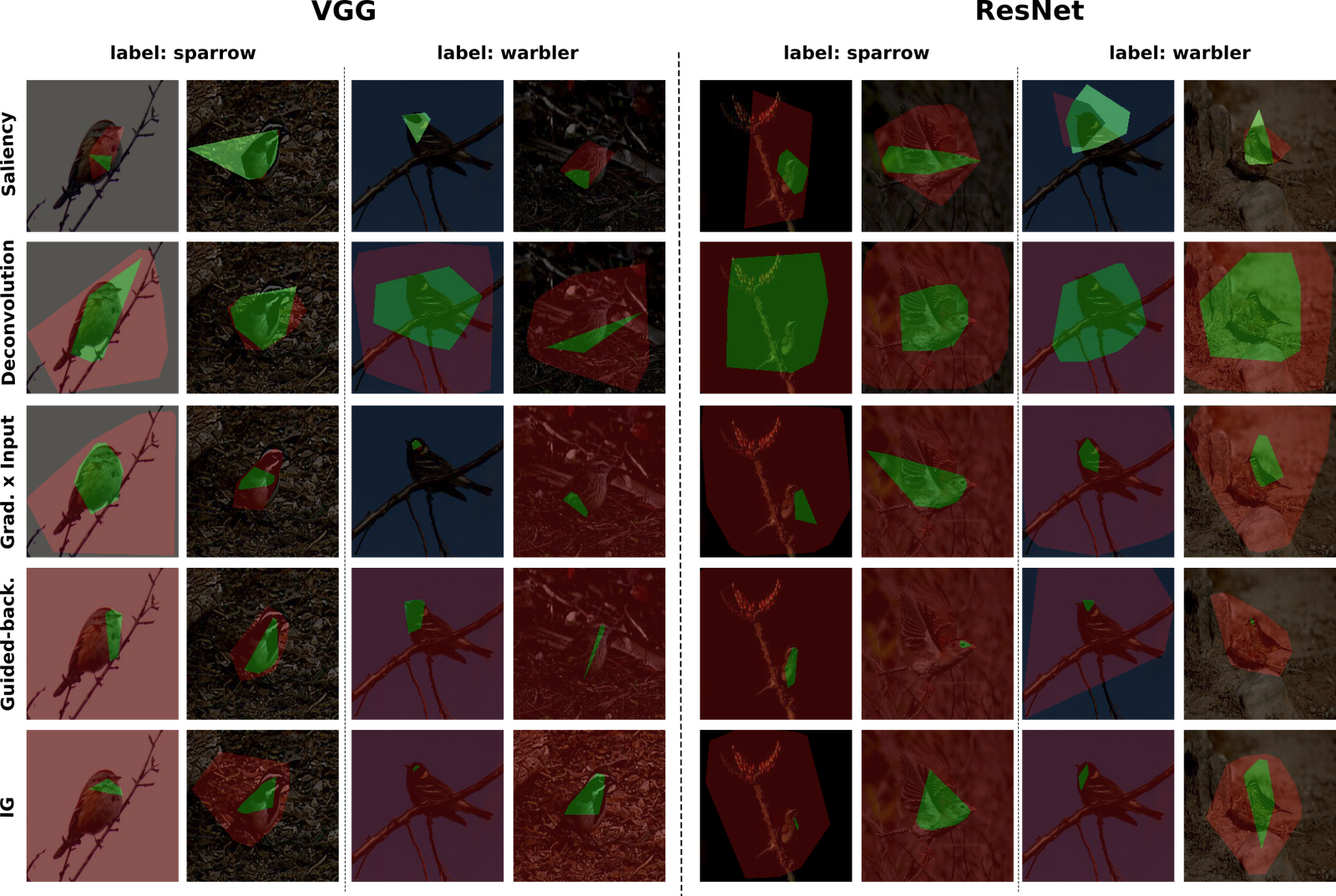} 
  \includegraphics[width=0.45\textwidth]{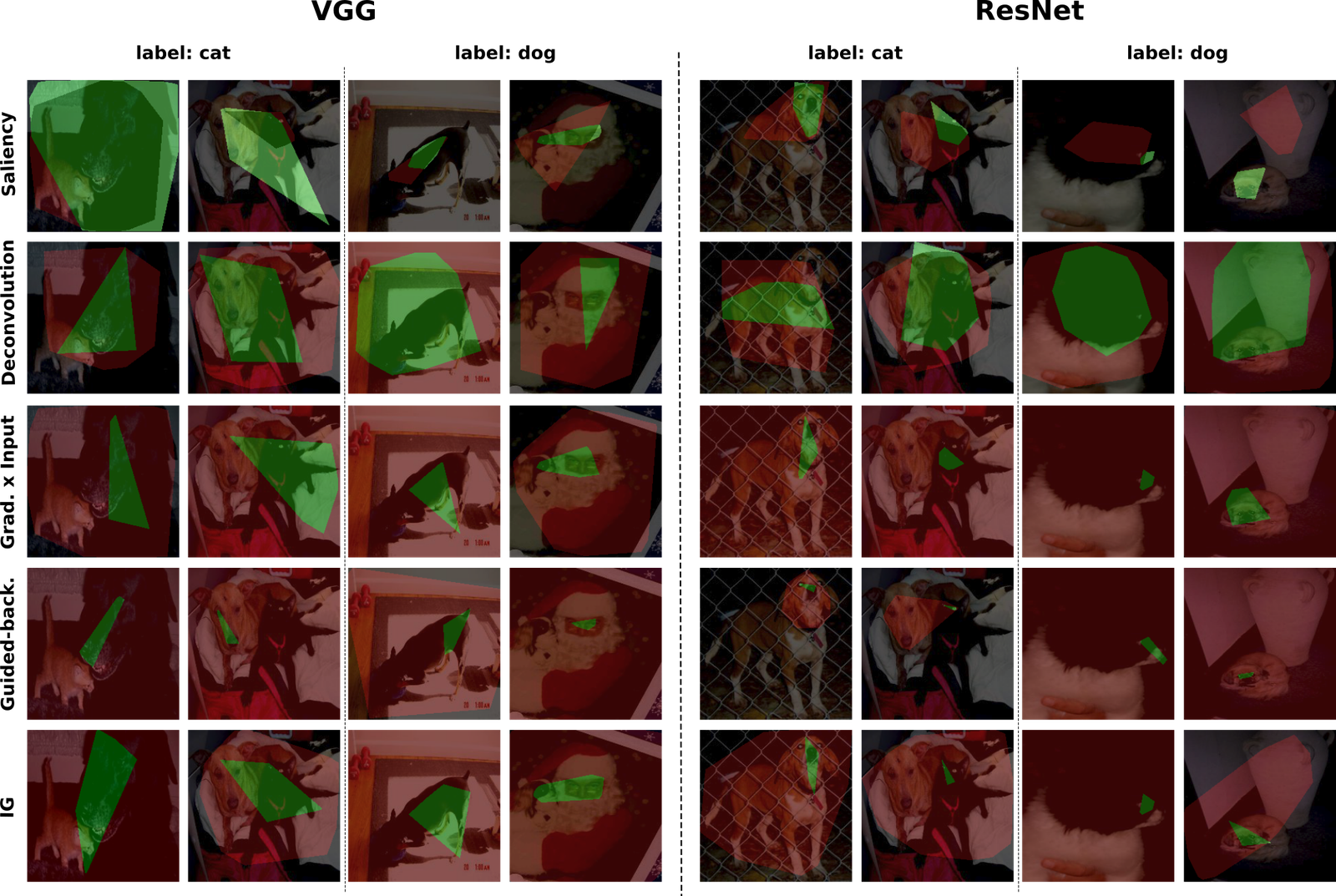} 
\caption{\textbf{Convex-hull comparisons for evaluation:} The convex-hull of \mymethodabbr's important features considerably reduces the images' interest regions. The comparison between original and \mymethodabbr convex hulls is able to give insights into why misclassifications happen. We show, in red, the original convex hull, and in green, the \mymethodabbr convex hull, for the five literature techniques (IG, Guided-Backpropagation, Gradient x Input, Deconvolution, and Saliency), the two architectures and the two datasets. We chose to visualize the less activated images for the corresponding label class in each model.}
\label{fig:all_datasets_analysis}
\end{figure*}

Figure~\ref{fig:all_datasets_analysis} illustrates a comparative visualization between original and GAD-interest areas. Aligned with the metrics detailed in Section~\ref{sec:eval_metrics}, we generated the convex hulls involving the most important pixels per visualization. We show the five less activated images for each assigned class (original label), for VGG (on the right) and ResNet (on the left), within the bird dataset (at the top), and the cat vs. dog dataset (at the bottom). In these visualizations, the GAD-obtained regions are highlighted in green, while the original regions are marked in red. We expect small green regions.

Additionally, we aimed to comprehend why these images were the least activated for their respective classes. While the bird dataset displays consistent behavior, the cat vs. dog dataset contains some images featuring both animals and occasionally people. Specifically, for VGG, the initial image showcases both animal types, with the dog (as per the larger green area) deemed more important than the cat. The final image features both a dog and Santa Claus, where both the animal and the person are deemed important. Regarding ResNet, the last image is intriguing as the dog is correctly identified as an important region, but the vase's presence likely diminishes the dog's activation for the dog class. In both models, the second-worst image is the same, containing both animal types, exemplifying differences between the models: VGG appears to recognize parts of both animal types, whereas ResNet emphasizes the cat, despite the dog's prominence impacting the cat's activation.

\subsection{\revOne{Extension for a multi-class problem}}

\begin{figure*}[!h]
\centering
 \includegraphics[width=0.9\textwidth]{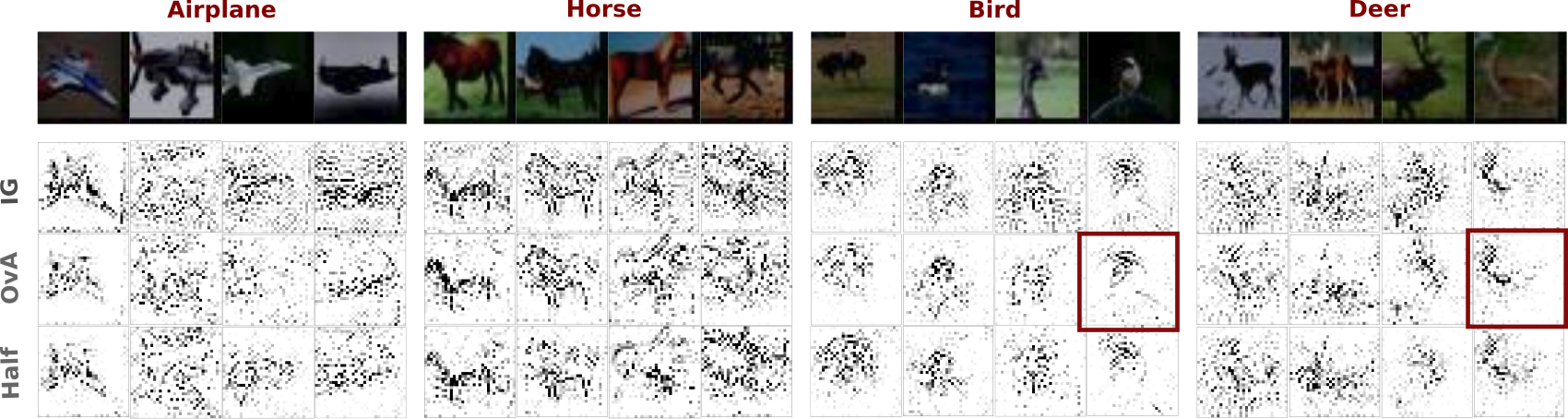} 
\caption{\revOne{Examples of the multi-class \mymethodabbr extension applied to Integrated Gradients (IG). We showcase the results of two strategies: One-vs-All (\textbf{OvA}) and Split output space (\textbf{Half}) for four classes from the CIFAR-10 dataset, using a VGG16 trained model. We observe a reduction in noise for both strategies, with smaller regions of interest highlighted by \textbf{OvA}.}}
\label{fig:multiclass}
\end{figure*}

\revOne{To test the multi-class extension idea presented in Section~\ref{sec:multiclass}, we employed the VGG architecture trained on the CIFAR-10 dataset \cite{krizhevsky:2009:cifar, krizhevsky:cifar} and Integrated Gradients (IG) explanation as base method. Results of the \textbf{OvA} and \textbf{Half} adaptations are illustrated in Figure~\ref{fig:multiclass}.}

\revOne{We observed two key points: firstly, both adaptations (\textbf{OvA} and \textbf{Half}) effectively reduce noise in the explanations. Secondly, \textbf{OvA} appears to yield more discernible features from the class with diminished regions of interest, as evidenced by the highlighted bird and deer in red.} 

\subsection{\revOne{Discussion of the findings}}

\revOne{Reducing the sparsity of the most important visualized areas enhances interpretability, enabling humans to focus on smaller, more informative regions. Our experiments (Figures~\ref{fig:cat_dog_attributions} and \ref{fig:multiclass}) demonstrate that \mymethodabbr effectively minimizes noise in explanation visualizations and highlights less sparse regions of interest. These areas are vital for the model's performance, as evidenced by the \textit{Sensitivity} experiment (Table~\ref{tab:sensitivity_all}). Moreover, our \textit{Complexity} experiment (Table~\ref{tab:complexity} and Figure~\ref{fig:all_datasets_analysis}) underscores \mymethodabbr's ability to produce concise explanations. Future work will involve utilizing density metrics, such as density histograms, and human-based evaluations to further validate interpretability quality metrics.}

%% file: conclusion.tex
\section{Conclusion}
\label{sec:conclusion}

The experiments demonstrated \mymethodabbr's efficacy in pinpointing the key regions influencing class relationships. Through mask-based occlusions, \mymethodabbr effectively highlighted pivotal areas in decision-making, even when dealing with smaller significant regions. Across datasets and models, \mymethodabbr reduced complexity and enhanced the sensitivity of five gradient-based explainability techniques. These findings deepened our interpretation of the networks' knowledge and provided insights into the underlying causes of certain misclassification errors. \revOne{Experiments in multi-class classification underscored the potential of \mymethodabbr in enhancing explanations and reducing noise. Overall, \mymethodabbr offers a straightforward approach to improving the interpretability of gradient-based methods for both binary and multi-class models by minimizing noise and directing attention to specific areas of interest. However, the method has its limitations, its interpretability hinges on human analysis. Consequently, if there is an overlap of important patterns, it may not be clear how to interpret and prioritize them. In future work, we aim to explore the automation of values for $\alpha$, additional metrics for interpreting quality, such as density measures and human-based evaluations, and a mechanism for clustering similar concepts (account for semantics) in different images to mitigate ambiguity of interpretation arising from overlapping important patterns.}